\documentclass[a4paper,fleqn]{cas-dc}

\usepackage[authoryear,round]{natbib}
\setcitestyle{aysep={,},citesep={; }} % (Du et al., 2019; Jin et al., 2022)

\usepackage{subcaption}

\definecolor{stdgray}{gray}{0.5}
\newcommand{\pmstd}[1]{{\scriptsize\textcolor{stdgray}{\,$\pm$\,#1}}}
\def\tsc#1{\csdef{#1}{\textsc{\lowercase{#1}}\xspace}}
\tsc{WGM}
\tsc{QE}
\begin{document}
\let\WriteBookmarks\relax
\def\floatpagepagefraction{1}
\def\textpagefraction{.001}

\shorttitle{}

% Short author (appears in the first-page footer before "Preprint submitted to Elsevier")
\shortauthors{de Jong et al.}

\title [mode = title]{ATLAS: A Multi-Procedure Benchmark and Context-Aware Model for Surgical Anatomy Segmentation}  

\tnotemark[1] 



\author[1]{Ronald L. P. D. {de Jong}}[orcid=0009-0005-7806-4340]

\cormark[1]

\fnmark[1]

\ead{r.l.p.d.d.jong@tue.nl}


\credit{Conceptualization, Data curation, Formal analysis, Investigation, Methodology, Software, Visualization, Writing – original draft, Writing – review \& editing}

\author[2]{Tim J. M. Jaspers}
\credit{Conceptualization, Data curation, Formal analysis, Investigation, Methodology, Software, Writing – review \& editing}

\author[2]{Raf A. H. Vervoort}
\credit{Data curation, Writing – review \& editing}

\author[3,4]{Aron F. H. A. Bakker}
\credit{Data curation, Writing – review \& editing}

\author[5]{Anna L. Dijk}
\credit{Data curation, Writing – review \& editing}

\author[1]{Yiping Li}
\credit{Methodology, Writing – review \& editing}

\author[6]{Jip L. Tolenaar}
\credit{Data curation, Writing – review \& editing}

\author[7]{Jelle P. Ruurda}
\credit{Data curation, Funding acquisition, Project administration, Writing – review \& editing}

\author[3]{Willem M. Brinkman}
\credit{Data curation, Writing – review \& editing}

\author[2]{Daan {de Geus}}
\credit{Methodology, Writing – review \& editing}

\author[2]{Fons {van der Sommen}}
\credit{Funding acquisition, Supervision, Writing – review \& editing}

\author[1]{Josien P. W. Pluim}
\credit{Funding acquisition, Project administration, Supervision, Writing – review \& editing}

\author[1,2]{Marcel Breeuwer}
\credit{Project administration, Supervision, Writing – review \& editing}

\affiliation[1]{organization={Department of Biomedical Engineering, Eindhoven
University of Technology},
            addressline={Groene Loper 3}, 
            postcode={5612 AE},
            postcodesep={},
            city={Eindhoven},
            %state={Noord-Brabant},
            country={The Netherlands}}

\affiliation[2]{organization={Department of Electrical Engineering, Eindhoven
University of Technology},
            addressline={Groene Loper 3}, 
            postcode={5612 AE},
            postcodesep={},
            city={Eindhoven},
            %state={Noord-Brabant},
            country={The Netherlands}}

\affiliation[3]{organization={Department of Oncological Urology, University Medical Center Utrecht},
            addressline={Heidelberglaan 100}, 
            postcode={3584 CX},
            postcodesep={},
            city={Utrecht},
            %state={Utrecht},
            country={The Netherlands}}

\affiliation[4]{organization={Department of Urology, Catharina Hospital},
            addressline={Michelangelolaan 2}, 
            postcode={5623 EJ},
            postcodesep={},
            city={Eindhoven},
            %state={Noord-Brabant},
            country={The Netherlands}}

\affiliation[5]{organization={Department of Technical Medicine, Delft University of Technology},
            addressline={Mekelweg 2}, 
            postcode={2628 CD},
            postcodesep={},
            city={Delft},
            %state={Zuid-Holland},
            country={The Netherlands}}

\affiliation[6]{organization={Department of Surgery, Haaglanden Medical Center},
            addressline={Lijnbaan 32}, 
            postcode={2512 VA},
            postcodesep={},
            city={The Hague},
            %state={Zuid-Holland},
            country={The Netherlands}}

\affiliation[7]{organization={Department of Surgery, University Medical Center Utrecht},
            addressline={Heidelberglaan 100}, 
            postcode={3584 CX},
            postcodesep={},
            city={Utrecht},
            %state={Utrecht},
            country={The Netherlands}}

\cortext[1]{Corresponding author}
\fntext[fn1]{First author}



\begin{abstract}
Automated recognition of anatomy in surgical video could support surgeons during minimally invasive and robot-assisted surgery, but existing anatomy segmentation datasets cover only a single procedure and few videos, and existing video segmentation models do not use the procedural context on which the presence of anatomical structures depends. We present ATLAS-120k, a dataset of 121,018 densely annotated frames from 502 clips of 100 videos, covering 14 laparoscopic and robot-assisted procedures and 41 classes. We also propose ATLAS, a video segmentation model that extends a query-based vision transformer with procedure and phase context queries, which are processed together with the segmentation queries and propagated between frames. The phase queries are trained with a self-supervised contrastive loss and require no phase labels. ATLAS outperforms natural-image and in-domain foundation models on both segmentation accuracy and temporal consistency, also at smaller model sizes, and runs well above real-time frame rates. Ablations show that query propagation mainly improves temporal consistency, while the phase queries mainly improve accuracy and boundary placement. ATLAS also generalizes beyond the dataset: applied zero-shot to CholecSeg8k, it matches a model trained on that dataset, and ATLAS pretraining improves surgical phase recognition on AutoLaparo. The ATLAS-120k annotations, annotation tool and model weights are publicly available at \url{https://github.com/TimJaspers0801/ATLAS}.
\end{abstract}

\begin{abstract}
Recognizing anatomical structures during minimally invasive surgery (MIS) could support intraoperative navigation and improve patient safety, but progress is limited by small, single-procedure datasets and by segmentation methods developed for natural scenes. This work addresses both limitations through a new dataset and model. First, we introduce ATLAS-120k, a clip-level semantic segmentation dataset of 121,018 annotated frames from 100 surgical videos, covering 14 laparoscopic and robot-assisted procedures and 41 classes. Annotations were created with a pipeline that combines manual labeling, tracker-based propagation, and review by surgeons. Second, we propose ATLAS (\underline{A}natomy Recognition with Contex\underline{t} \underline{L}earning using Found\underline{a}tion Representation\underline{s}), a video semantic segmentation model for surgical anatomy recognition. ATLAS builds on in-domain foundation-model embeddings and adds context queries for the procedure and surgical phase, together with query propagation between frames for short-term memory. ATLAS outperforms state-of-the-art natural-image and in-domain foundation models in both segmentation accuracy and temporal consistency, achieving a Dice score of 0.70 with ViT-L while running in real time at 76.2 FPS. Ablations show that the context queries improve accuracy and that query propagation improves temporal consistency. ATLAS also generalizes well, matching a dedicated model zero-shot on CholecSeg8k and improving phase recognition on AutoLaparo over SurgeNet initialization. The model weights, dataset annotations, and annotation tool are publicly available at \url{https://github.com/TimJaspers0801/ATLAS}.
\end{abstract}


% \begin{highlights}
% \item 121k annotated frames from 100 videos covering 14 laparoscopic and robotic procedures
% \item Dense labels for 41 anatomical and instrument classes, with inter-rater analysis
% \item ATLAS conditions segmentation on learned procedure and phase context queries
% \item ATLAS outperforms foundation model baselines in accuracy and temporal consistency
% \item Annotations, model weights and the annotation tool are publicly released
% \end{highlights}

%\nocite{*}

% Keywords

\maketitle

% Main text
\section{Introduction}\label{sec:introduction}Intraoperative surgical computer vision holds potential to transform minimally invasive surgery (MIS) and robot-assisted surgery (RAS) by delivering real-time decision support, automated anatomical navigation, and enhanced patient safety. However, translating computer vision algorithms into dependable clinical guidance systems remains challenging. In the absence of haptic feedback and under constrained endoscopic fields of view, surgeons must continually interpret complex anatomical relationships while navigating adverse visual conditions such as tissue deformation, bleeding, and surgical smoke. Achieving robust clinical utility requires automated vision models capable of dense, real-time anatomical perception that generalizes reliably across diverse surgical environments~\citep{MAIERHEIN2022, denBoer2023, Mascagni2022}.

Despite significant progress in surgical scene understanding, prior research has largely prioritized constrained sub-tasks, including surgical phase recognition~\citep{heichole, Cholec80}, instrument detection and tracking~\citep{surgloc}, and tool segmentation~\citep{roboticinstrumentsegmentation}. While these tasks provide foundational capabilities for surgical workflow analysis, they do not offer the dense, semantic spatial reasoning required for anatomy-guided navigation. Existing anatomy segmentation benchmarks have taken initial steps toward bridging this gap, but suffer from key systemic limitations: they are typically restricted to a single surgical procedure (most notably laparoscopic cholecystectomy), encompass small patient cohorts with limited video counts, or offer only isolated, single-organ segmentations rather than dense, holistic scene annotations~\citep{dsad, Cholecseg8k, Mascagni2025}. Consequently, models trained on current benchmarks struggle to generalize across varying anatomical topologies, surgical modalities, and procedural contexts.

Methodologically, existing video semantic segmentation frameworks adapted from natural image domains fall short in surgical settings because they treat video frames as generic visual sequences without exploiting intraoperative structural domain priors. In clinical practice, expert surgical cognition is inherently hierarchical and context-driven: a surgeon's expectation of class presence, anatomical exposure, and structural boundary location is conditioned on the overarching surgical procedure type and the immediate surgical phase. Standard memory- or tracking-based segmentation networks prioritize localized frame-to-frame pixel correspondence while failing to model these global temporal and procedural priors.

To address these bottlenecks, we introduce a unified dataset and algorithmic framework designed to establish a scalable foundation for anatomy-aware surgical perception. Our main contributions are summarized as follows:

\begin{itemize}

\item \textbf{ATLAS-120k Dataset:} We present a novel, large-scale surgical video dataset comprising $121,018$ densely annotated frames across $502$ video clips sourced from $100$ surgical procedures. Spanning $14$ distinct laparoscopic and robot-assisted MIS operations and $41$ semantic classes, ATLAS-120k expands procedural, technological, and anatomical diversity compared to existing benchmarks.

\item \textbf{ATLAS Framework:} We propose ATLAS (\underline{A}natomy Recognition with Contex\underline{t} \underline{L}earning using Found\underline{a}tion Representation\underline{s}), a video semantic segmentation model specifically tailored to surgical scene understanding. ATLAS combines strong visual embeddings from a surgical foundation model~\citep{jaspers2025surgenet} with explicit procedure- and phase-level context queries, enabling global context-conditioned feature learning alongside short-term query propagation for temporally stable predictions.

\item \textbf{Open resources and insights:} We publicly release the ATLAS-120k annotations and the ATLAS model weights. We also report on the dataset creation process, including inter-observer variability and how well tracker-based propagation reduces annotation effort, and analyze the model through ablations of its components, training clip length and amount of annotated data, per-class results, and visualizations of its predictions.

\end{itemize}

This work extends our MICCAI conference paper~\citep{dejong2026surgical}. Compared with that version, we add a review of related work, a more detailed description and analysis of the dataset, including its feature-space diversity, an extended five-step annotation protocol with full-sequence validation, an inter-observer variability study, and an analysis of tracker-based annotation propagation. The ablation study is extended with leave-one-out experiments for each component and analyses of training clip length and the amount of annotated data. We further report per-class results and inference speed for all model sizes, and evaluate generalization through zero-shot transfer to CholecSeg8k and transfer to surgical phase recognition on AutoLaparo.

\section{Related work}

\begin{figure*}[t]
    \centering
    \includegraphics[width=0.99\linewidth]{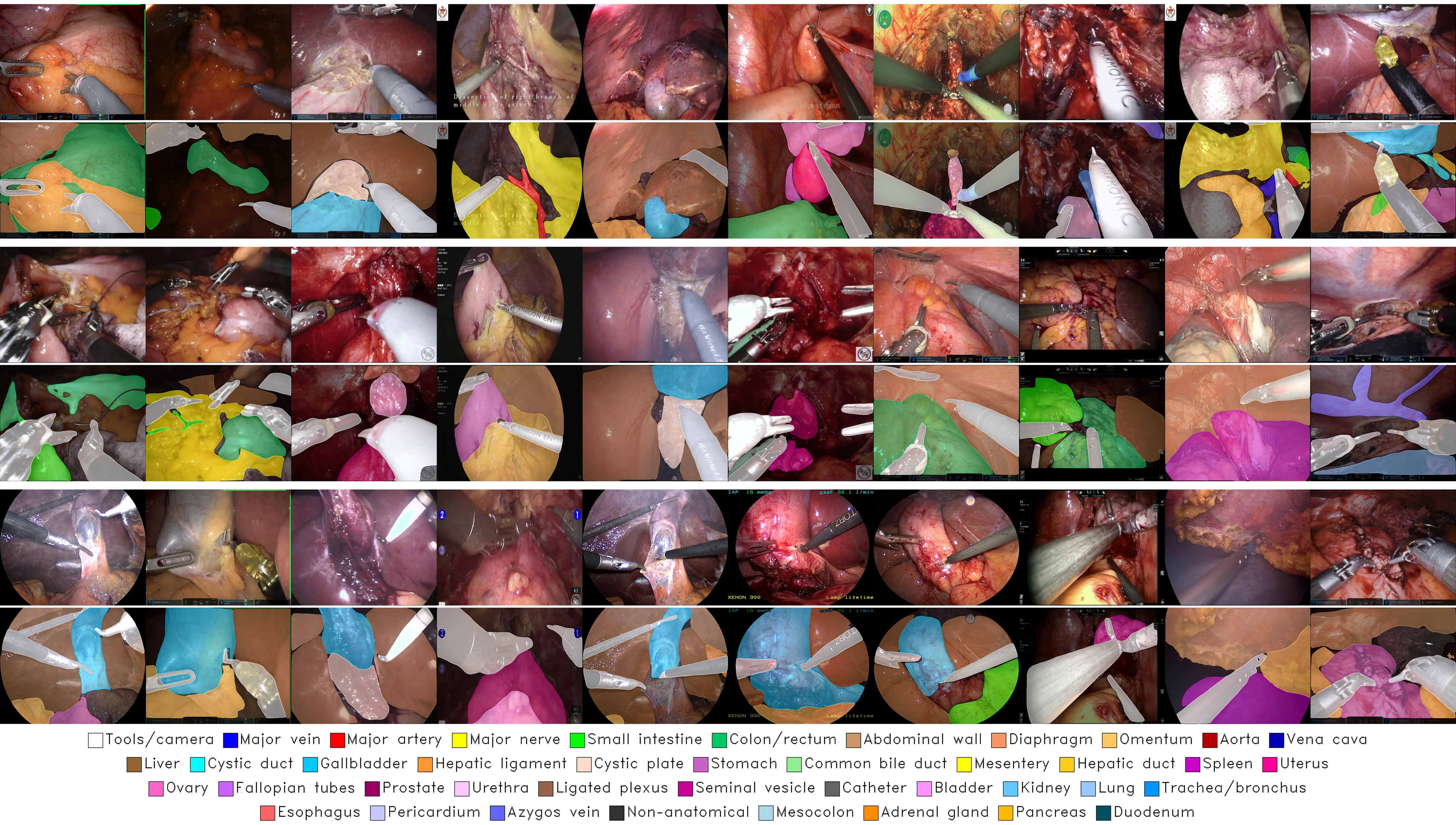}
    \caption{Randomly selected frames from ATLAS-120k. Odd rows show the original frames and even rows the same frames with their annotations overlaid. Each of the 14 procedures and each semantic class is represented in at least one example.}
    \label{fig:examples}
\end{figure*}

\subsection{Surgical anatomy segmentation datasets}
Several public datasets have advanced anatomy-oriented segmentation in minimally invasive surgery (MIS), but each remains constrained in scope. CholecSeg8k~\citep{Cholecseg8k} provides 8,080 frames drawn from 17 laparoscopic cholecystectomy videos of the Cholec80 dataset, annotated with 12 classes covering anatomy and instruments. The Dresden Surgical Anatomy Dataset (DSAD)~\citep{dsad} offers a larger pool of 13,195 frames across 32 videos with 11 anatomical classes; however, the majority of its frames depict isolated organ segmentations rather than dense, full-scene semantic labels, which limits its suitability for holistic scene understanding. Endoscapes-Seg50~\citep{Mascagni2025} targets Critical-View-of-Safety assessment during cholecystectomy, providing 493 annotated frames drawn from 50 videos across 6 classes. Collectively, these resources share limitations that constrain the generalizability of models trained on them: each covers only a single surgical procedure, so representations learned from them may not generalize well to other procedural contexts; the number of source videos is modest (17-50); and anatomical coverage is narrow (6-12 classes). \text{ATLAS-120k} addresses these limitations by covering multiple procedures, both laparoscopic and robot-assisted, with dense full-scene annotations in a single taxonomy. Table~\ref{tab:surgnetseg75k} compares \text{ATLAS-120k} with existing datasets.

\begin{table*}[]
    \setlength{\tabcolsep}{4pt}
    \centering
    \caption{Comparison of surgical video segmentation dataset characteristics. MIS indicates minimally invasive surgery; RA indicates robot-assisted procedures.}
    \begin{tabular}{l | c c c c c c c }
        \hline
        \toprule
       \textbf{Dataset}           & \textbf{Included}  & \textbf{\#} & \textbf{\#}  & \textbf{\#} & \textbf{\#}  & \textbf{MIS}  & \textbf{RA} \\
       & \textbf{procedures} & \textbf{classes} & \textbf{videos} & \textbf{clips} & \textbf{frames} & & \\
        \midrule
        Endoscapes-Seg50~\citep{Mascagni2025}  & 1                    & 6          & 50          & -        & 493 & \checkmark &  \\
        CholecSeg8k~\citep{Cholecseg8k}       & 1                    & 12         & 17          & 101        & 8,080 & \checkmark & \\   
        DSAD~\citep{dsad}              & 1                    & 11         & 32          & -       & 13,195 &  & \checkmark \\   
        ATLAS-120k    & 14                   & 41         & 100         & 502       & 121,018 & \checkmark & \checkmark \\
        \bottomrule
    \end{tabular}
    \label{tab:surgnetseg75k}
\end{table*}

\subsection{Procedural and phase context in surgical video understanding}
A separate line of work has shown that information about the overall procedure provides useful context for interpreting surgical video. EndoNet~\citep{Cholec80} was among the first to show that tool presence and surgical phase could be learned jointly from a shared CNN backbone, indicating that these two aspects of surgical context are entangled and mutually informative. TeCNO~\citep{tecno} reformulated phase recognition as a temporal segmentation problem, using a multi-stage temporal convolutional network with causal, dilated convolutions over pre-extracted frame features to obtain a large receptive field while retaining online, real-time inference. Trans-SVNet~\citep{transsvnet} subsequently showed that fusing spatial (frame-level) and temporal (phase-level) embeddings through a lightweight cross-attention transformer improves phase recognition accuracy over purely temporal models, highlighting the complementarity of instantaneous visual evidence and longer-range procedural context. A common limitation of current work is that phase and procedure recognition are treated as separate tasks, rather than being used to inform downstream perception such as segmentation. ATLAS builds on the value of procedural context by incorporating procedure and phase information directly into the segmentation model through additional query tokens. These context queries are learned jointly with the segmentation task, allowing information about the current procedural context to influence the anatomical predictions.

\subsection{Foundation models for surgical and endoscopic vision}
Motivated by the success of large-scale self-supervised pretraining on natural images, such as DINOv2~\citep{2024dinov2} and DINOv3~\citep{simeoni2025dinov3}, a growing body of work adapts this paradigm to the surgical and endoscopic domain to compensate for the relative scarcity of densely annotated data. EndoViT~\citep{EndoVit} pretrains a masked-autoencoder vision transformer on a large corpus of endoscopic images, while EndoFM~\citep{endofm} instead performs spatio-temporal self-supervised pretraining directly on unlabeled endoscopic video. GastroNet-5M~\citep{GastroNet5m} scales this approach further using a multi-center gastrointestinal endoscopy dataset. In the surgical domain specifically, SurgeNet and its large-scale successor SurgeNetXL~\citep{jaspers2025surgenet} systematically scale self-supervised pretraining for surgical video, while GSViT~\citep{schmidgall2024general} and LEMON~\citep{che2025lemon} pursue related video-level pretraining strategies for general surgery and endoscopy, respectively. Collectively, these models demonstrate that domain-specific pretraining yields embeddings that transfer more effectively to downstream surgical perception tasks than their natural-image counterparts. ATLAS builds directly on this line of work: rather than training a new foundation encoder from scratch, it leverages strong embeddings from a surgical foundation model together with in-domain pretrained weights~\citep{jong2026towards}, and instead focuses its contribution on the lightweight temporal and contextual reasoning required to exploit these embeddings for consistent, procedure-aware video segmentation.

\subsection{Video segmentation and temporal memory}
Most video segmentation models used in surgery address video object segmentation, where the targets are given in the first frame and tracked without class labels. Cutie~\citep{cheng2023putting} and SAM2~\citep{ravi2025sam2} store features and masks of past frames in an explicit memory bank. Surgical adaptations of SAM2 make this memory more robust to occlusion and smoke~\citep{yin2025memory} or suppress predictions when the target is out of view~\citep{liu2026uni}. These methods need a prompt per target and do not predict anatomical classes. SAM3~\citep{carion2026sam} replaces object prompts with concept prompts, but is still queried one concept at a time and was trained on general-domain concepts.

Our setting is video semantic segmentation, where every pixel is assigned one of a fixed set of classes. VidEoMT~\citep{videomt} addresses this task with a plain vision transformer that processes learned queries together with the image patches, following EoMT~\citep{eomt}, so no separate decoder is needed. Each query predicts a class and a mask. The output queries of each frame are passed on to the next frame, so a query keeps representing the same structure over time at a constant cost per frame. What is carried forward, however, is limited to the tracked structures. VidEoMT has no representation of the procedure or of how far it has progressed. ATLAS builds on VidEoMT and adds this context (Section~\ref{sec:atlas_model}).

\section{Methods}

\subsection{ATLAS-120k: Anatomy Segmentation Dataset}

\subsubsection{Data Sourcing and Processing}
A total of 100 videos were sourced from the open-source GSViT dataset \citep{schmidgall2024general}, from which 14 distinct procedures were selected for annotation. Rather than annotating full videos, clips of varying lengths were extracted to maximize the breadth of scenes. To standardize the dataset, all videos were downsampled to 15 fps, which corresponds to the lowest original frame rate observed across the collection. Spatial resolutions across the videos varied from 480 $\times$ 640 to 1080 $\times$ 1920 pixels. Figure \ref{fig:examples} shows representative examples from the ATLAS-120k dataset, with at least one example per procedure. 

\begin{figure*}[]
    \centering
    \includegraphics[width=0.90\linewidth]{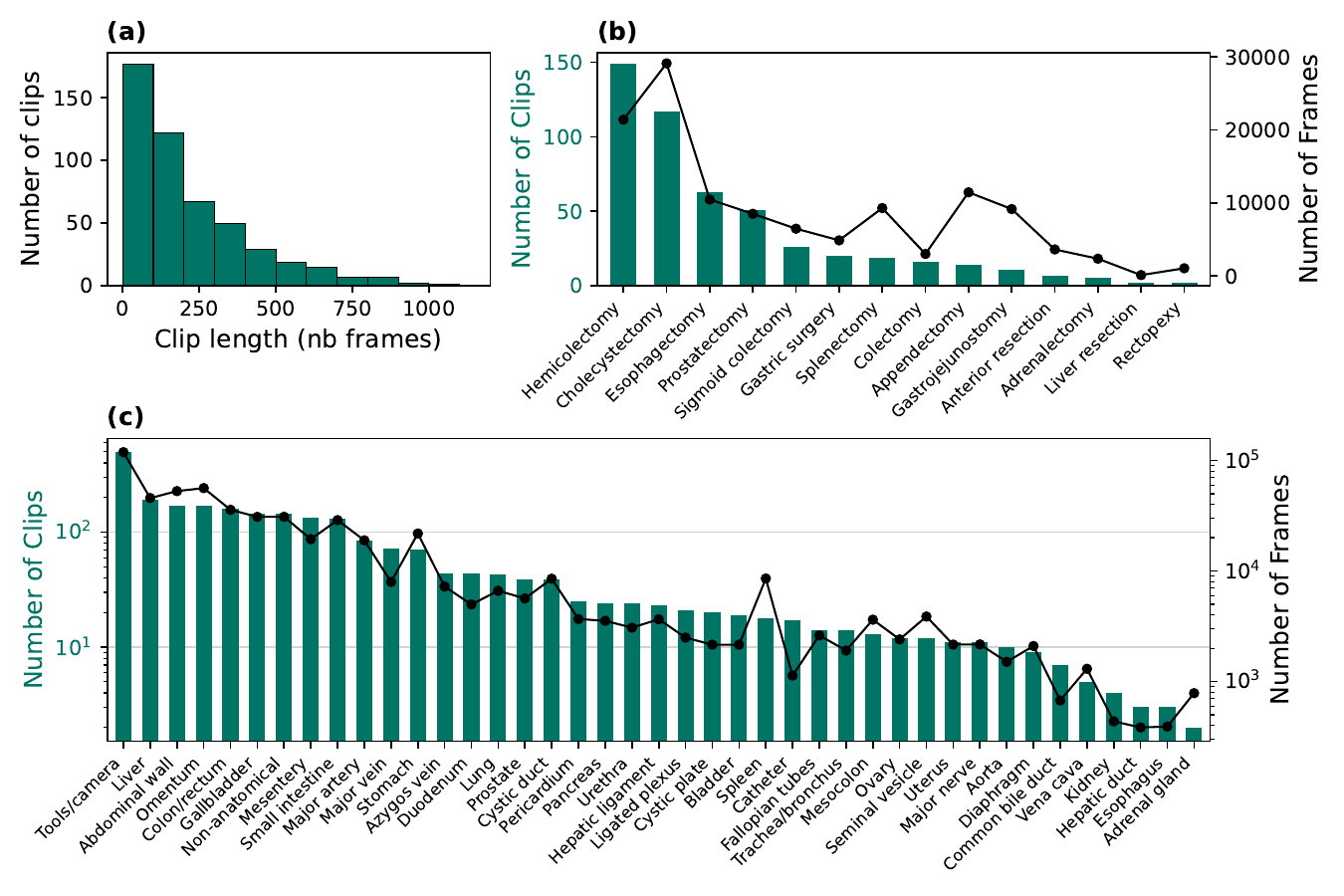}
    \caption{ATLAS-120k dataset characteristics.}
    \label{fig:data_characteristics}
\end{figure*}

\subsubsection{Diversity and Class Distribution}
\begin{figure*}[h]
    \centering
    \includegraphics[width=0.99\linewidth]{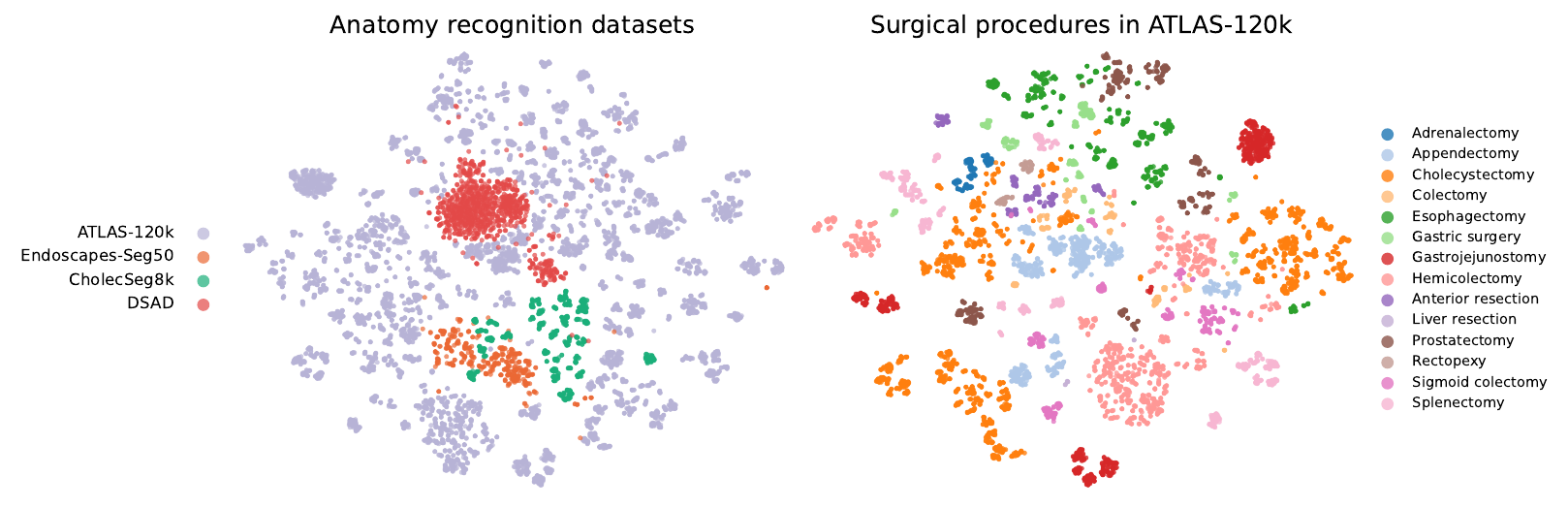}
    \caption{t-SNE projection of SurgeNet~\citep{jaspers2025surgenet} frame embeddings; each point is one frame, and frames with similar features lie close together. Left: ATLAS-120k compared with Endoscapes-Seg50, CholecSeg8k and DSAD, colored by dataset. Right: ATLAS-120k colored by procedure.}
    \label{fig:tsne}
\end{figure*}

A core contribution of ATLAS-120k is its procedural and anatomical diversity, as detailed across three dimensions in Figure~\ref{fig:data_characteristics}. First, Figure~\ref{fig:data_characteristics}a presents the distribution of clip durations. Most clips contain fewer than 400 frames (under 30 seconds), as we extracted many short clips per video to cover a wide range of scenes rather than annotating full videos. Second, Figure~\ref{fig:data_characteristics}b illustrates the number of clips and frames per procedure. This distribution is uneven, reflecting the availability of each procedure in the GSViT dataset~\citep{schmidgall2024general}. Third, Figure~\ref{fig:data_characteristics}c highlights the clips and frames per semantic class, clearly demonstrating a long-tail class distribution. This inherent class imbalance reflects the reality of diverse intraoperative environments and introduces a critical benchmark challenge for models to reliably recognize rare but essential anatomical structures. To further illustrate this diversity, Figure~\ref{fig:tsne} shows a t-SNE projection of frame embeddings extracted with a frozen SurgeNet backbone~\citep{jaspers2025surgenet}. The left pane contrasts ATLAS-120k against related benchmarks (Endoscapes-Seg50, CholecSeg8k, and DSAD), illustrating its broader feature space. The right pane visualizes the internal diversity across the 14 distinct surgical procedures within ATLAS-120k, highlighting the substantial inter-procedural variations that models must be equipped to navigate.

\subsubsection{Annotation Protocol and Quality Assurance}
Annotations were produced by two surgical research fellows under the direct supervision of three experienced surgeons, each possessing over 10 years of clinical expertise. Although the primary objective was anatomical structure recognition, all visible surgical instruments were also annotated to enable additional applications such as surgical tool segmentation. The annotation workflow was conducted in five controlled steps to ensure data integrity:

\begin{enumerate}
    \item \textbf{Manual Initialization:} The initial frame of each extracted clip was manually annotated with high precision.
    \item \textbf{Propagation and Refinement:} A lightweight object-tracking model \citep{cheng2023putting} was used to propagate these initial annotations across all subsequent frames. When propagation errors occurred, the affected frame was manually corrected and subsequently used as a new initialization point to resume the automated tracking. To improve tracking performance and efficiency during this phase, the model was fine-tuned on all clips annotated so far. This was repeated once approximately 10\%, 25\%, 50\%, and 100\% of the clips had been annotated, using the original training parameters of \citep{cheng2023putting}.
    \item \textbf{Expert Review:} The first frame of every clip was independently reviewed by at least one experienced surgeon to verify the clinical correctness and completeness of the anatomical labels.
    \item \textbf{Comprehensive Sequence Validation:} The full clip was comprehensively reviewed by an independent expert rater to guarantee high annotation quality and temporal consistency across the sequence.
    \item \textbf{Inter-rater agreement analysis:} To assess annotation reliability, the test set (see Section~\ref{sec:dataset_split}) was additionally annotated by a second, independent annotator.
\end{enumerate}

The dataset annotations are released as open source to actively facilitate further research in this direction.

\subsection{ATLAS: Segmentation Model} \label{sec:atlas_model}
Many anatomical structures in surgical video are difficult to identify from appearance alone. Adjacent tissues often have a similar color and texture, and structures are partly covered by fat or connective tissue, deformed by retraction, or obscured by blood and smoke. Surgeons partly resolve this ambiguity with context. Knowing which procedure is being performed and how far it has progressed, they know which structures to expect and roughly where to find them. ATLAS is designed to give a segmentation model access to this type of context, while keeping its predictions consistent from frame to frame.

ATLAS builds on VidEoMT~\citep{videomt}, a video segmentation model based on a plain vision transformer (ViT). A ViT divides each frame into patches of $16 \times 16$ pixels and represents each patch as a token, the patch tokens. VidEoMT adds a set of learnable query tokens to this sequence, and the ViT processes the patch and query tokens together. At the output, each query predicts a class label and a mask for one structure in the frame. To handle video, the output queries of frame $t-1$ are used as input queries for frame $t$. A query that segmented a structure in one frame therefore starts the next frame with what it learned about that structure, so the model can keep track of structures it has already identified, and its predictions stay consistent over time.

We use VidEoMT as a starting point for two reasons. First, because it represents what it segments as query tokens, context can be added as extra queries without changing the architecture. The segmentation queries can then attend to these context queries inside the same encoder. Second, each frame is processed only once and only the queries are carried forward, so the cost per frame stays constant and the model runs in real time.

\begin{figure*}[t]
    \centering
    \includegraphics[width=0.85\linewidth]{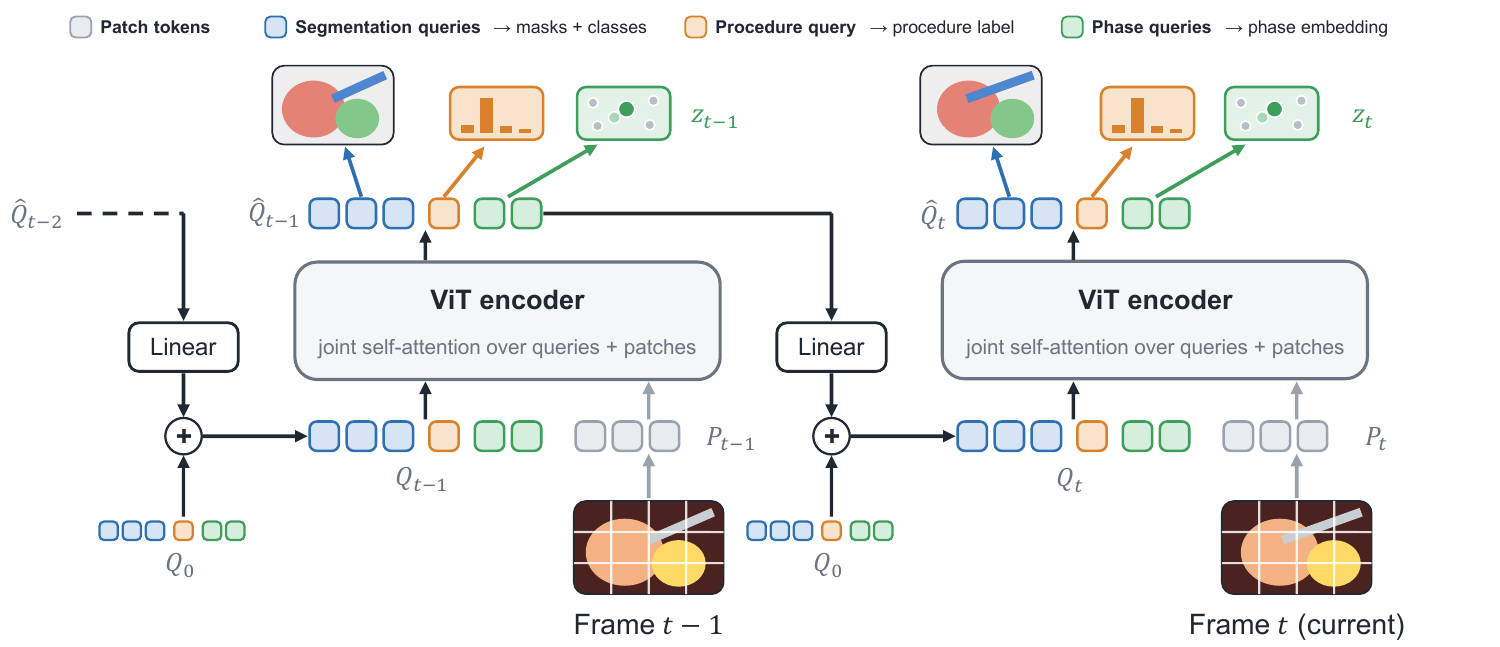}
    \caption{Overview of ATLAS. For each frame, a ViT encoder jointly processes the patch tokens $P_t$ and the query tokens $Q_t$, which consist of segmentation, procedure and phase queries. The output queries $\hat{Q}_t$ predict the masks and classes, the procedure label and a phase embedding $z_t$. They are then propagated to the next frame as $Q_{t+1} = \mathrm{Linear}(\hat{Q}_t) + Q_0$. Only past frames are used, so the model runs online. This schematic was created with the assistance of Claude (Anthropic).}
    \label{fig:method}
\end{figure*}

The information VidEoMT carries between frames describes only the individual structures it tracks. It has no representation of which procedure is being performed or how far it has progressed. ATLAS adds this information through two types of context queries, a procedure query and phase queries. They are processed in the same encoder as the segmentation queries, propagated to the next frame in the same way, and trained with their own auxiliary losses. Figure~\ref{fig:method} gives an overview of the model.

\subsubsection{Transformer Input}
At frame $t$, the encoder receives the token sequence
\begin{equation}
    X_t = (P_t,\; Q^\textrm{seg}_{t},\; Q^\textrm{proc}_{t},\; Q^\textrm{phase}_{t} ),
\end{equation}
where $P_t$ denotes the visual patch tokens of the current frame and $Q^\textrm{seg}_{t}$, $Q^\textrm{proc}_{t}$ and $Q^\textrm{phase}_{t}$ the segmentation, procedure and phase queries. The query tokens are built from two sources. The initial queries $Q^\textrm{seg}_{0}$, $Q^\textrm{proc}_{0}$ and $Q^\textrm{phase}_{0}$ are learnable parameters shared across all frames and videos, and act as a learned prior for each query type. The propagated queries $\hat{Q}^\textrm{seg}_{t-1}$, $\hat{Q}^\textrm{proc}_{t-1}$ and $\hat{Q}^\textrm{phase}_{t-1}$ are the output queries of the previous frame and carry what the model has observed so far: the segmentation queries hold information about individual structures, and the context queries hold information about the procedure and its progression. The input queries at frame $t$ are obtained by mapping the propagated queries with a linear layer and adding the initial queries. All tokens are processed by a single transformer encoder. Because self-attention is global, each segmentation query can attend both to the visual evidence in $P_t$ and to the context queries within the same layer. The context therefore shapes the class and mask predictions directly, rather than being fused with them at a later stage.

\subsubsection{Procedure Context Queries}
The procedure query represents information that holds for an entire video: the type of operation being performed. At every frame, the output procedure query is mapped by a multilayer perceptron (MLP) to predict a distribution of scores over all possible procedure classes,
\begin{equation}
\hat y^\textrm{proc}_t = \text{MLP}(\hat Q^\textrm{proc}_t),
\end{equation}
which is compared with the ground-truth label $y^\textrm{proc}_t$ using the cross-entropy loss
\begin{equation}
\mathcal{L}_\textrm{proc} = \text{CE}(\hat y^\textrm{proc}_t, y^\textrm{proc}_t).
\end{equation}
Procedure classification is not a goal in itself. The loss serves to make the query summarize the global cues that identify a procedure, such as the instruments in use and the organs in view. Since the segmentation queries attend to this summary, they can use it to judge which anatomical classes are plausible in the current scene.

\subsubsection{Phase Context Queries}
Within a procedure, the expected anatomy also changes over time, as dissection gradually exposes new structures. The phase queries are intended to capture this progression. Rather than supervising them with phase labels, we train them in a self-supervised manner. For each frame, the output phase queries are averaged and projected by an MLP to an $\ell_2$-normalized embedding
\begin{equation}
z_t = \text{MLP}(\hat Q^\textrm{phase}_t).
\end{equation}
These embeddings are optimized with a contrastive loss. Each training batch consists of short windows of consecutive frames, and the other frames in the same window as frame $t$ form its set of positives $P(t)$. Since each frame has several positives, we use the multi-positive form of the InfoNCE loss~\citep{khosla2020supervised}:
\begin{equation}
\mathcal{L}_\textrm{phase} = -\frac{1}{|P(t)|} \sum_{t^+ \in P(t)} \log \frac{\exp(z_t \cdot z_{t^+}/\tau)}{\sum_{j \neq t} \exp(z_t \cdot z_j/\tau)},
\end{equation}
where $j$ runs over all other frames in the batch and $\tau$ is a temperature parameter. The loss is averaged over all frames in the batch. Frames from other windows serve as negatives. The loss thus pulls together the embeddings of frames from the same window and pushes apart those of different windows, which typically come from different procedures or from different moments within a procedure. As a result, the phase queries learn a representation of the surgical state that varies smoothly within a clip and distinguishes between different stages of a procedure.

\subsubsection{Temporal Query Propagation}
Following VidEoMT~\citep{videomt}, queries are passed from one frame to the next to provide short-term memory and temporal consistency. The input queries $Q_t$ at frame $t$ are obtained from the output queries $\hat{Q}_{t-1}$ of the previous frame and the initial learnable queries $Q_0$:
\begin{equation}
Q_t = \mathrm{Linear}(\hat{Q}_{t-1}) + Q_0,
\end{equation}
where $\mathrm{Linear}$ is a learned linear layer and
\begin{equation}
\begin{aligned}
\hat{Q}_{t-1} &= 
\begin{bmatrix}
\hat{Q}^\textrm{seg}_{t-1} & \hat{Q}^\textrm{proc}_{t-1} & \hat{Q}^\textrm{phase}_{t-1}
\end{bmatrix}, \\
\text{and} \quad Q_0 &= 
\begin{bmatrix}
Q^\textrm{seg}_0 & Q^\textrm{proc}_0 & Q^\textrm{phase}_0
\end{bmatrix}.
\end{aligned}
\end{equation}
The linear layer maps the output queries back to the input query space, and the added initial queries retain a learned prior in every query, so that the model can still detect structures that newly enter the field of view. Segmentation and context queries are propagated by the same mechanism. The segmentation queries thereby keep track of individual structures, while the context queries accumulate information about the procedure over time. Because each frame depends only on the queries of the previous frame, the model runs online, and its computational cost per frame does not grow with the length of the video.

\subsubsection{Loss function}
The segmentation loss is taken from EoMT~\citep{eomt}: masks are supervised with a binary cross-entropy loss $\mathcal{L}_{\text{bce}}$ and a Dice loss $\mathcal{L}_{\text{dice}}$, and class predictions with a cross-entropy loss $\mathcal{L}_{\text{ce}}$. Adding the procedure loss $\mathcal{L}_{\text{proc}}$ and the phase loss $\mathcal{L}_{\text{phase}}$ gives the total loss
\begin{equation}
\begin{split}
\mathcal{L}_{\text{total}} ={} & \lambda_{\text{bce}} \mathcal{L}_{\text{bce}} + \lambda_{\text{dice}} \mathcal{L}_{\text{dice}} + \lambda_{\text{ce}} \mathcal{L}_{\text{ce}} \\
& + \lambda_{\text{proc}} \mathcal{L}_{\text{proc}} + \lambda_{\text{phase}} \mathcal{L}_{\text{phase}},
\end{split}
\end{equation}
where the weights $\lambda$ balance the terms.

\subsubsection{Pretraining} 
The encoder is initialized with in-domain weights obtained by DINOv3~\citep{simeoni2025dinov3} pretraining on the SurgeNet dataset~\citep{jaspers2025surgenet}, following~\citep{jong2026towards}. This adapts the encoder to surgical scenes before training on ATLAS-120k. Because the encoder already captures the appearance of surgical anatomy and instruments, the model starts from features that are better suited to the target domain than natural-image features.

\section{Experimental setup}
\subsection{Dataset} \label{sec:dataset_split}
The ATLAS-120k dataset was partitioned into patient-level training (70 videos), validation (10 videos), and test (20 videos) splits. All 14 procedures were represented in the training set, while the validation and test sets covered 6 and 9 procedures, respectively. Model training was restricted to 27 classes. Specific anatomical classes were mapped to broader parent classes within a hierarchical label structure: vena cava and azygos vein to vein, aorta to artery, cystic duct, common bile duct, and hepatic duct to bile duct, and omentum and mesentery to fat. Catheter was mapped to non-anatomical. Classes absent from the training or test set, or with fewer than 500 annotated training frames (esophagus), were relabeled as background. The size and class distribution of each split are given in Appendix~\ref{app:splits}.

\subsection{ATLAS implementation}
ATLAS uses a DINOv3 ViT-S/16, ViT-B/16 or ViT-L/16 encoder at an input resolution of $256 \times 256$ pixels, with the queries inserted before the last four encoder blocks, as in EoMT~\citep{eomt}. We use 100 segmentation queries, one procedure query and four phase queries. The model is trained on windows of five consecutive frames with a batch size of 24 windows, for 10 epochs on a single NVIDIA H100 GPU, using AdamW (learning rate $10^{-4}$, weight decay 0.05, layer-wise decay 0.8), the warm-up and polynomial decay schedule of EoMT, and mixed precision. Augmentation consists of random flipping, scaling, cropping and color jitter, applied identically to all frames of a window. For the phase loss, temperature parameter $\tau$ was set to $0.07$. The loss weights $\lambda_{\text{bce}}$, $\lambda_{\text{dice}}$, $\lambda_{\text{ce}}$, $\lambda_{\text{proc}}$, and $\lambda_{\text{phase}}$ were set to 5.0, 5.0, 2.0, 1.0, and 0.1,
respectively. The checkpoint with the highest validation Dice is used for testing. Ablations use ViT-L, and all hyperparameters are kept fixed unless otherwise stated.

\subsection{Compared methods}
Baselines are initialized from their public pretrained weights and fine-tuned with a pixel-wise cross-entropy loss. DINOv2, DINOv3 and GastroNet5M use a linear head, the SurgeNet models an FPN decoder, EndoViT a DPT decoder, and EndoFM a TransUNet decoder; these encoders are fine-tuned end-to-end. SAM2-UNet and SAM3-UNet keep their image encoders frozen and train lightweight adapters together with a U-Net decoder, following~\cite{SAM2UNet}.

\subsection{Evaluation}
Models are evaluated on the ATLAS-120k test set, predicting frame by frame in temporal order. Segmentation accuracy is measured with Dice and 95th percentile Hausdorff distance (HD95). Both are first computed per structure in each frame, then averaged within each clip, and finally averaged over clips. HD95 is normalized by the image diagonal, and a class that is present but not predicted receives a value of 1. The per-class results use a pixel-based Dice, computed from pixel counts summed over the entire test set. This gives a more stable estimate for classes that occur in only a few clips or are small in many frames. Temporal consistency is measured with mVC$_{12}$ and mVC$_{24}$~\citep{miao2021vspw}: the fraction of non-background pixels with a constant ground-truth label over a sliding window of 12 or 24 frames whose prediction also stays constant. As this measures stability rather than correctness, we interpret it together with Dice and HD95.

\subsection{Generalization experiments}
Generalization is assessed on CholecSeg8k~\citep{Cholecseg8k} for segmentation and on AutoLaparo~\citep{wang2022auto} for phase recognition. For CholecSeg8k, we use videos 12, 20, 48 and 55 for testing, 1 and 18 for validation, and the remaining 11 for training, and merge grasper and L-hook into one instrument class, giving eight classes. The fine-tuned reference uses SurgeNet initialization. For zero-shot evaluation, the predictions of the unmodified ATLAS model are mapped to these classes: small intestine, colon/rectum and stomach to gastrointestinal tract; hepatic ligament and cystic plate to connective tissue; tools/camera to instrument; matching classes directly; and all others to background.

For phase recognition, the ATLAS ViT-L encoder initializes a causal temporal convolutional network~\citep{tecno} (four residual blocks, dilations 1--8) applied to frame features sampled at 1~fps and resized to $256 \times 256$. The model is trained end-to-end for 20 epochs with cross-entropy and AdamW (learning rate $10^{-4}$ for the head, $10^{-5}$ for the encoder), on 10 training, 4 validation and 7 test videos. We report frame-wise accuracy, balanced accuracy, F1 score and Jaccard index.

\subsection{Data leakage}
To prevent data leakage, we used the variant of SurgeNet pretrained without YouTube data, ensuring that ATLAS-120k videos were not included in pretraining. The training, validation, and test splits of ATLAS-120k were made at the video level, ensuring that no video appears in more than one split. For CholecSeg8k, all videos in the respective test sets were excluded from the pretraining data. AutoLaparo was not part of the pretraining data. Thus, all evaluations were conducted on videos not seen during pretraining or model training.

\section{Results}
\subsection{Annotation quality}
\subsubsection{Inter-observer variability}
To assess the reliability of the annotations, the full test set was annotated independently by a second annotator. Agreement was computed with the same metrics and per-clip averaging as the model evaluation, using the original annotation as reference and the second annotation as prediction. Table~\ref{tab:inter_observer} shows the results. The annotators reached a Dice of $0.85$ and an HD95 of $0.10$. The annotations were also temporally stable, with an mVC$_{12}$ of $0.96$ and an mVC$_{24}$ of $0.94$. ATLAS ViT-L remains below this level on all metrics. The gap is largest for Dice and mVC$_{24}$ (0.15 each), and smaller for mVC$_{12}$ (0.11) and HD95 (0.07). The model therefore falls behind the annotators both in segmentation accuracy and in temporal consistency over longer horizons.

\begin{table}
\centering
\small
\setlength{\tabcolsep}{3pt}
\caption{Inter-observer variability compared with ATLAS ViT-L on the full test set. Values are the mean $\pm$ standard deviation over clips.}
\label{tab:inter_observer}
\begin{tabular}{l | c c c c}
\toprule
 & Dice $\uparrow$ & HD95 $\downarrow$ & mVC12 $\uparrow$ & mVC24 $\uparrow$ \\
\midrule
Inter-observer & 0.85\pmstd{0.10} & 0.10\pmstd{0.08} & 0.96\pmstd{0.05} & 0.94\pmstd{0.08} \\
ATLAS ViT-L    & 0.70\pmstd{0.13} & 0.17\pmstd{0.11} & 0.85\pmstd{0.10} & 0.79\pmstd{0.15} \\
\bottomrule
\end{tabular}
\end{table}

\subsubsection{Annotation propagation performance}
\begin{figure}
    \centering
    \includegraphics[width=0.99\linewidth]{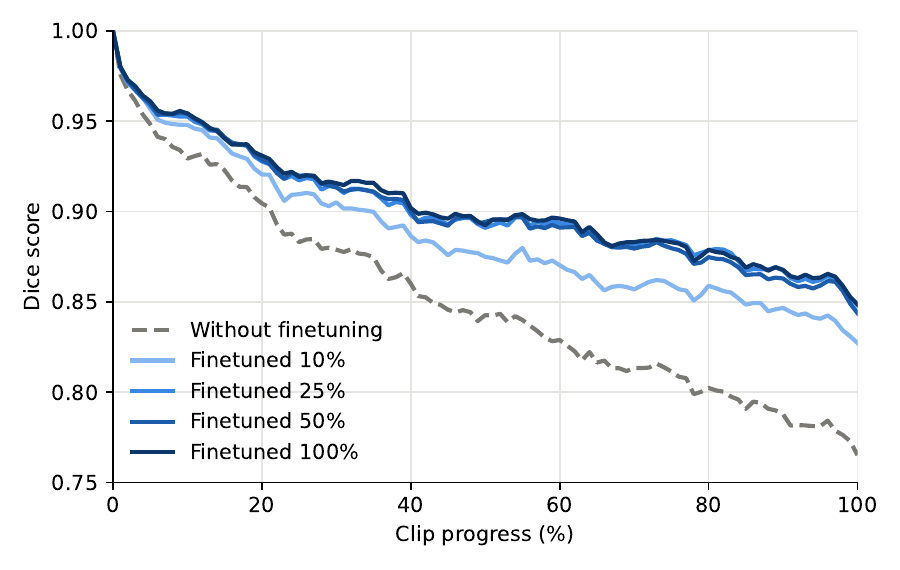}
    \caption{Dice score between annotations propagated from the first frame without manual correction and the final test-set annotations, as a function of the position in the clip. Results are shown for the tracker without fine-tuning and after fine-tuning on 10--100\% of the training set.}
    \label{fig:tracking}
\end{figure}

Figure~\ref{fig:tracking} shows how accurately the tracker in the annotation tool propagates an annotation over a clip. For each test clip, the manually annotated first frame was propagated to all subsequent frames without any correction. The result was compared with the final annotations using the Dice score. Results are shown for the tracker without fine-tuning and after fine-tuning on 10\%, 25\%, 50\%, and 100\% of the training set. For all trackers, Dice decreases steadily with the distance to the first frame. Without fine-tuning, it drops to approximately 0.84 halfway through the clip and to 0.77 at the end. The propagated annotations thus never match the final annotations exactly, so manual correction was needed throughout each clip. Fine-tuning improves propagation at every point in the clip. With 10\% of the training data, the final Dice increases to approximately 0.83. With 25\% or more, it reaches approximately 0.85, and the curves for 25\%, 50\%, and 100\% are almost identical. Most of the improvement from fine-tuning is thus obtained with 25\% of the training data.

\subsection{Segmentation performance and efficiency}
\subsubsection{Comparison with existing methods}
Table~\ref{tab:sota_results} compares ATLAS with natural-image foundation models and in-domain pretrained medical models on the ATLAS-120k test set. ATLAS ViT-L achieves the best results on all four metrics. ATLAS ViT-B, with 94M parameters, also outperforms all baselines on every metric. This includes SAM3-UNet, which has almost five times as many parameters (449M) and reaches a Dice of 0.60. Compared with the best baseline for each metric, ATLAS ViT-L improves Dice by 0.09, HD95 by 0.06, mVC12 by 0.07, and mVC24 by 0.09, so it improves both segmentation accuracy and temporal consistency.

At a matched backbone size, the comparison isolates the effect of the ATLAS architecture and pretraining. Among ViT-B models, ATLAS reaches a Dice of 0.67, against 0.57 for EndoViT and 0.56 for DINOv3. ATLAS ViT-S (24M parameters) reaches a Dice of 0.57, which matches EndoViT with roughly a fifth of the parameters. However, it remains below the three SurgeNet models (0.60--0.61 Dice, 26--30M parameters) on all metrics. At the smallest scale, these hierarchical backbones therefore remain competitive, and the advantage of ATLAS emerges from ViT-B onwards.

Among the natural-image models, DINOv3 outperforms DINOv2 (0.56 vs.\ 0.49 Dice for ViT-B). For both models, scaling from ViT-B to ViT-L does not improve results, which suggests that a linear head limits how well these embeddings can be used for dense surgical segmentation. In-domain pretraining helps only in some cases. EndoViT and the three SurgeNet models are among the strongest baselines. In contrast, EndoFM and GastroNet5M fall below DINOv3, even though both were pretrained on endoscopic data. ATLAS is the only model that ranks first on both accuracy and consistency.

\begin{table*}[t]
\setlength{\tabcolsep}{4pt}
\centering
\caption{Quantitative comparison of foundation models for segmentation. Models are grouped by pretraining domain: natural-image foundation models (top), in-domain pretrained medical models (middle), and our proposed ATLAS models (bottom). Values are the mean $\pm$ standard deviation over test clips.
For DINOv2 and DINOv3, we train a linear layer attached to the ViT. Cell shading is scaled linearly within each column from the worst (white) to the best (dark green) value. $^{*}$Statistically significant improvement over the best baseline for that metric, i.e., the best-performing non-ATLAS model ($p < 0.01$, Wilcoxon signed-rank test on paired per-clip scores).}
\begin{tabular}{l l c | c c c c}
\toprule
Model & Backbone & \#PARAMS (M) $\downarrow$ & Dice $\uparrow$ & HD95 $\downarrow$ & mVC12 $\uparrow$ & mVC24 $\uparrow$ \\
\midrule
DINOv2~\citep{2024dinov2} & ViT-B
& \cellcolor{green!30}87
& \cellcolor{green!13}0.49\pmstd{0.14}
& \cellcolor{green!10}0.32\pmstd{0.13}
& \cellcolor{green!19}0.70\pmstd{0.13}
& \cellcolor{green!17}0.59\pmstd{0.18} \\

DINOv2~\citep{2024dinov2} & ViT-L
& \cellcolor{green!12}304
& \cellcolor{green!0}0.36\pmstd{0.12}
& \cellcolor{green!0}0.38\pmstd{0.12}
& \cellcolor{green!6}0.58\pmstd{0.15}
& \cellcolor{green!6}0.46\pmstd{0.19} \\

DINOv3~\citep{simeoni2025dinov3} & ViT-B
& \cellcolor{green!30}86
& \cellcolor{green!21}0.56\pmstd{0.14}
& \cellcolor{green!20}0.26\pmstd{0.12}
& \cellcolor{green!22}0.73\pmstd{0.14}
& \cellcolor{green!22}0.64\pmstd{0.18} \\

DINOv3~\citep{simeoni2025dinov3} & ViT-L
& \cellcolor{green!12}303
& \cellcolor{green!20}0.55\pmstd{0.14}
& \cellcolor{green!20}0.26\pmstd{0.11}
& \cellcolor{green!21}0.72\pmstd{0.14}
& \cellcolor{green!21}0.63\pmstd{0.19} \\

SAM2-UNet~\citep{SAM2UNet} & Hiera-L
& \cellcolor{green!19}217
& \cellcolor{green!19}0.54\pmstd{0.14}
& \cellcolor{green!18}0.27\pmstd{0.11}
& \cellcolor{green!15}0.66\pmstd{0.17}
& \cellcolor{green!14}0.55\pmstd{0.21} \\

SAM3-UNet~\citep{SAM3UNet} & Hiera-L
& \cellcolor{green!0}449
& \cellcolor{green!25}0.60\pmstd{0.14}
& \cellcolor{green!23}0.24\pmstd{0.11}
& \cellcolor{green!25}0.76\pmstd{0.14}
& \cellcolor{green!25}0.67\pmstd{0.20} \\

\midrule
EndoFM~\citep{endofm} & ViT-B
& \cellcolor{green!25}141
& \cellcolor{green!1}0.37\pmstd{0.12}
& \cellcolor{green!0}0.38\pmstd{0.13}
& \cellcolor{green!0}0.52\pmstd{0.15}
& \cellcolor{green!0}0.39\pmstd{0.18} \\

GastroNet5M~\citep{GastroNet5m} & ViT-B
& \cellcolor{green!30}86
& \cellcolor{green!9}0.45\pmstd{0.14}
& \cellcolor{green!8}0.33\pmstd{0.13}
& \cellcolor{green!16}0.67\pmstd{0.15}
& \cellcolor{green!14}0.55\pmstd{0.20} \\

EndoViT~\citep{EndoVit} & ViT-B
& \cellcolor{green!28}112
& \cellcolor{green!22}0.57\pmstd{0.16}
& \cellcolor{green!18}0.27\pmstd{0.12}
& \cellcolor{green!22}0.73\pmstd{0.16}
& \cellcolor{green!22}0.64\pmstd{0.21} \\

SurgeNet-pvt~\citep{jaspers2025surgenet} & PVTv2
& \cellcolor{green!35}27
& \cellcolor{green!25}0.60\pmstd{0.14}
& \cellcolor{green!25}0.23\pmstd{0.09}
& \cellcolor{green!28}0.78\pmstd{0.14}
& \cellcolor{green!27}0.70\pmstd{0.19} \\

SurgeNet-conv~\citep{jaspers2025surgenet} & ConvNeXtv2
& \cellcolor{green!35}30
& \cellcolor{green!25}0.60\pmstd{0.14}
& \cellcolor{green!23}0.24\pmstd{0.10}
& \cellcolor{green!25}0.76\pmstd{0.14}
& \cellcolor{green!25}0.67\pmstd{0.19} \\

SurgeNet-ca~\citep{jaspers2025surgenet} & CaFormer-S18
& \cellcolor{green!35}26
& \cellcolor{green!26}0.61\pmstd{0.14}
& \cellcolor{green!23}0.24\pmstd{0.11}
& \cellcolor{green!27}0.77\pmstd{0.14}
& \cellcolor{green!26}0.69\pmstd{0.19} \\

\midrule
ATLAS (ours) & ViT-S
& \cellcolor{green!35}24
& \cellcolor{green!22}0.57\pmstd{0.14}
& \cellcolor{green!13}0.30\pmstd{0.12}
& \cellcolor{green!22}0.73\pmstd{0.14}
& \cellcolor{green!22}0.64\pmstd{0.18} \\

ATLAS (ours) & ViT-B
& \cellcolor{green!29}94
& \cellcolor{green!32}0.67\pmstd{0.13}
& \cellcolor{green!28}0.21\pmstd{0.11}
& \cellcolor{green!33}0.83\pmstd{0.11}
& \cellcolor{green!33}0.77\pmstd{0.16} \\

ATLAS (ours) & ViT-L
& \cellcolor{green!11}318
& \cellcolor{green!35}\textbf{0.70}$^{*}$\pmstd{0.13}
& \cellcolor{green!35}\textbf{0.17}$^{*}$\pmstd{0.11}
& \cellcolor{green!35}\textbf{0.85}$^{*}$\pmstd{0.10}
& \cellcolor{green!35}\textbf{0.79}$^{*}$\pmstd{0.15} \\

\bottomrule
\end{tabular}
\label{tab:sota_results}
\end{table*}

\subsubsection{Inference speed}
Table~\ref{tab:inference_speed} reports the inference speed of ATLAS on a single NVIDIA H100 GPU. ATLAS ViT-S runs at 263.7 FPS (3.79~ms per frame), ATLAS ViT-B at 160.3 FPS (6.24~ms per frame), and ATLAS ViT-L at 76.2 FPS (13.16~ms per frame). Because the model processes each frame once and carries information forward only through the propagated queries, the cost per frame does not grow with the length of the video. All three variants therefore run well above the frame rate of the ATLAS-120k videos (15 FPS) and of typical endoscopic video (25--30 FPS), making online use during surgery feasible.

\subsection{Qualitative results}
\begin{figure*}
    \centering
    \includegraphics[width=0.9\linewidth]{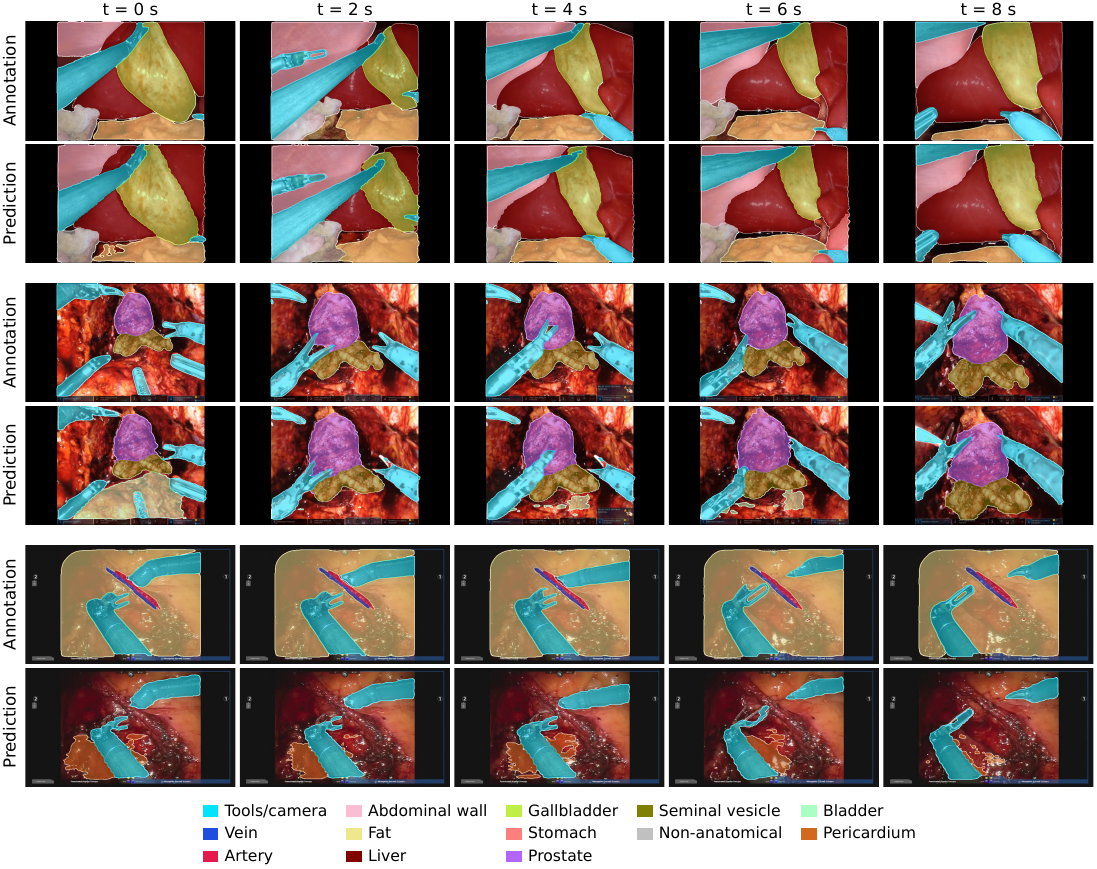}
    \caption{Qualitative segmentation results of ATLAS ViT-L on three test clips: robotic cholecystectomy (top), robot-assisted radical prostatectomy (RARP, middle) and robotic hemicolectomy (bottom). Each example shows five frames sampled every 2 s from the same clip.}
    \label{fig:qualitative}
\end{figure*}

Figure~\ref{fig:qualitative} shows predictions of ATLAS ViT-L on three test clips. In the cholecystectomy and RARP examples, the predictions closely follow the annotations. Instruments, the liver, gallbladder and abdominal wall, and the prostate and seminal vesicles are segmented with accurate boundaries, and their masks remain stable over the 8~seconds shown, including when instruments move across or partly occlude them. The remaining errors are mostly local: fat is segmented in fragments, the seminal vesicle boundary varies slightly between frames, and small false-positive regions appear. The hemicolectomy example shows a clear failure case. The instruments are segmented correctly, but the thin artery and vein are missed in all frames, and most of the fat is not recognized.

\begin{table}[t]
\centering
\small
\setlength{\tabcolsep}{4pt}
\caption{Inference speed of ATLAS on a single NVIDIA H100 GPU.}
\label{tab:inference_speed}
\begin{tabular}{l | c c c}
\toprule
Model & \#Params (M) & Latency (ms) $\downarrow$ & FPS $\uparrow$ \\
\midrule
ATLAS ViT-S & 23.8  & 3.79  & 263.7 \\
ATLAS ViT-B & 94.3  & 6.24  & 160.3 \\
ATLAS ViT-L & 318.3 & 13.16 & 76.2 \\
\bottomrule
\end{tabular}
\end{table}

\subsection{Per-class results}
Table~\ref{tab:results_per_class} reports the results of ATLAS ViT-L for each class. Dice ranges from 0.94 for tools/camera to 0.03 for the hepatic ligament, and 13 of the 27 classes reach 0.70 or higher. The best-segmented classes are instruments and large organs with clear boundaries, such as the lung (0.91), liver (0.89), and gallbladder (0.86), which also have a low HD95 (0.03--0.11). The lowest Dice is found for thin or tubular structures, namely arteries (0.35), bile ducts (0.33), airways (0.22), and nerves (0.17), and for the spleen (0.17) and hepatic ligament (0.03). The latter two also have the highest HD95 (0.84 and 0.81), which indicates that they are often missed entirely. Temporal consistency generally follows segmentation accuracy: large organs such as the liver and uterus reach an mVC12 of 0.87 and 0.91, while arteries and nerves reach 0.52 and 0.55. There are several exceptions. The ovary and the non-anatomical class combine a relatively high Dice (0.74 and 0.83) with low consistency (0.44 and 0.57 mVC12), so their predictions are mostly correct but fluctuate over time. Conversely, the hepatic ligament and ligated plexus have higher consistency (0.69 and 0.90 mVC12) than their Dice would suggest. For the hepatic ligament, this reflects predictions that are stable but incorrect. Per-class consistency for rare classes is based on few test clips and should be interpreted with care.

\begin{table}[!htbp]
\centering
\small
\setlength{\tabcolsep}{5pt}
\caption{Per-class results of ATLAS ViT-L on the ATLAS-120k test set, sorted by pooled Dice. Pooled Dice is computed over all test frames together; HD95 is averaged over frames; mVC12 and mVC24 are averaged over the test clips in which the class occurs. Cell shading is scaled linearly within each column from the worst (white) to the best (dark green) value.}
\label{tab:results_per_class}
\begin{tabular}{l c c c c}
\toprule
Class & Dice $\uparrow$ & HD95 $\downarrow$ & mVC12 $\uparrow$ & mVC24 $\uparrow$ \\
\midrule
Tools/camera     & \cellcolor{green!35}0.94 & \cellcolor{green!35}0.03 & \cellcolor{green!32}0.87 & \cellcolor{green!31}0.81 \\
Lung             & \cellcolor{green!34}0.91 & \cellcolor{green!35}0.04 & \cellcolor{green!29}0.83 & \cellcolor{green!29}0.77 \\
Liver            & \cellcolor{green!33}0.89 & \cellcolor{green!32}0.11 & \cellcolor{green!32}0.87 & \cellcolor{green!33}0.83 \\
Gallbladder      & \cellcolor{green!32}0.86 & \cellcolor{green!33}0.07 & \cellcolor{green!30}0.84 & \cellcolor{green!30}0.79 \\
Non-anatomical   & \cellcolor{green!31}0.83 & \cellcolor{green!29}0.18 & \cellcolor{green!10}0.57 & \cellcolor{green!11}0.47 \\
Ovary            & \cellcolor{green!27}0.74 & \cellcolor{green!29}0.18 & \cellcolor{green!0}0.44  & \cellcolor{green!0}0.29 \\
Pericardium      & \cellcolor{green!27}0.73 & \cellcolor{green!30}0.14 & \cellcolor{green!12}0.60 & \cellcolor{green!13}0.50 \\
Fat              & \cellcolor{green!27}0.72 & \cellcolor{green!26}0.24 & \cellcolor{green!29}0.83 & \cellcolor{green!30}0.79 \\
Vein             & \cellcolor{green!27}0.72 & \cellcolor{green!24}0.28 & \cellcolor{green!27}0.80 & \cellcolor{green!27}0.73 \\
Abdominal wall   & \cellcolor{green!26}0.71 & \cellcolor{green!28}0.20 & \cellcolor{green!31}0.85 & \cellcolor{green!31}0.80 \\
Uterus           & \cellcolor{green!26}0.71 & \cellcolor{green!24}0.29 & \cellcolor{green!35}0.91 & \cellcolor{green!35}0.87 \\
Seminal vesicle  & \cellcolor{green!26}0.70 & \cellcolor{green!32}0.10 & \cellcolor{green!32}0.87 & \cellcolor{green!33}0.83 \\
Prostate         & \cellcolor{green!26}0.70 & \cellcolor{green!29}0.18 & \cellcolor{green!28}0.82 & \cellcolor{green!26}0.72 \\
Cystic plate     & \cellcolor{green!23}0.63 & \cellcolor{green!32}0.10 & \cellcolor{green!10}0.58 & \cellcolor{green!9}0.44 \\
Colon/rectum     & \cellcolor{green!22}0.59 & \cellcolor{green!21}0.35 & \cellcolor{green!18}0.68 & \cellcolor{green!16}0.55 \\
Ligated plexus   & \cellcolor{green!20}0.56 & \cellcolor{green!34}0.05 & \cellcolor{green!34}0.90 & \cellcolor{green!34}0.86 \\
Urethra          & \cellcolor{green!17}0.46 & \cellcolor{green!28}0.19 & \cellcolor{green!8}0.55  & \cellcolor{green!8}0.42 \\
Small intestine  & \cellcolor{green!15}0.43 & \cellcolor{green!19}0.40 & \cellcolor{green!16}0.66 & \cellcolor{green!15}0.54 \\
Fallopian tubes  & \cellcolor{green!14}0.40 & \cellcolor{green!29}0.18 & \cellcolor{green!18}0.68 & \cellcolor{green!16}0.55 \\
Artery           & \cellcolor{green!12}0.35 & \cellcolor{green!22}0.33 & \cellcolor{green!6}0.52  & \cellcolor{green!7}0.40 \\
Stomach          & \cellcolor{green!12}0.34 & \cellcolor{green!18}0.43 & \cellcolor{green!12}0.60 & \cellcolor{green!16}0.55 \\
Bladder          & \cellcolor{green!12}0.33 & \cellcolor{green!7}0.68  & \cellcolor{green!16}0.65 & \cellcolor{green!16}0.56 \\
Bile duct        & \cellcolor{green!12}0.33 & \cellcolor{green!25}0.27 & \cellcolor{green!10}0.58 & \cellcolor{green!7}0.41 \\
Trachea/bronchus & \cellcolor{green!7}0.22  & \cellcolor{green!17}0.45 & \cellcolor{green!26}0.79 & \cellcolor{green!25}0.70 \\
Major nerve      & \cellcolor{green!5}0.17  & \cellcolor{green!34}0.05 & \cellcolor{green!8}0.55  & \cellcolor{green!6}0.39 \\
Spleen           & \cellcolor{green!5}0.17  & \cellcolor{green!0}0.84  & \cellcolor{green!10}0.58 & \cellcolor{green!11}0.47 \\
Hepatic ligament & \cellcolor{green!0}0.03  & \cellcolor{green!1}0.81  & \cellcolor{green!19}0.69 & \cellcolor{green!19}0.60 \\
\bottomrule
\end{tabular}
\end{table}

\subsection{Ablation studies}
\subsubsection{Contribution of model components}
Table~\ref{tab:model_components} shows the contribution of each ATLAS component. Removing in-domain pretraining causes the largest overall drop, lowering Dice from 0.70 to 0.67 and both consistency metrics by 0.03. A strong surgical initialization therefore benefits both accuracy and temporal stability. The other components each mainly affect a different metric. Without temporal propagation, Dice and HD95 change little, but mVC12 and mVC24 drop to the level of the model without in-domain pretraining. Query propagation is therefore the main source of temporal consistency. Removing the phase queries has the opposite effect: consistency is unchanged, but Dice drops to 0.67 and HD95 increases from 0.17 to 0.23, the largest boundary error of all variants. This suggests that phase context mainly helps the model decide which structures are present and where they are, rather than keeping predictions stable over time. Removing the procedure queries gives a smaller drop in accuracy (0.69 Dice, 0.19 HD95) with little effect on consistency. Because each component improves a different aspect of performance, only the full model reaches the best accuracy and consistency together.

\begin{table}[h]
\centering
\small
\setlength{\tabcolsep}{3pt}
\caption{Ablation studies showing the effect of in-domain pretraining, temporal propagation, and procedure and phase queries on surgical video segmentation. Values are the mean $\pm$ standard deviation over test clips. Cell shading is scaled linearly within each column from the worst (white) to the best (dark green) value.}
\label{tab:model_components}
\resizebox{\columnwidth}{!}{%
\begin{tabular}{l | c c c c}
\toprule
Method & Dice $\uparrow$ & HD95 $\downarrow$ & mVC12 $\uparrow$ & mVC24 $\uparrow$ \\
\midrule
Full
& \cellcolor{green!35}0.70\pmstd{0.13}
& \cellcolor{green!35}0.17\pmstd{0.11}
& \cellcolor{green!31}0.85\pmstd{0.10}
& \cellcolor{green!31}0.79\pmstd{0.15} \\

w/o in-domain pretr.
& \cellcolor{green!0}0.67\pmstd{0.13}
& \cellcolor{green!17}0.20\pmstd{0.10}
& \cellcolor{green!1}0.82\pmstd{0.11}
& \cellcolor{green!0}0.76\pmstd{0.15} \\

w/o temporal prop.
& \cellcolor{green!22}0.69\pmstd{0.13}
& \cellcolor{green!27}0.18\pmstd{0.10}
& \cellcolor{green!0}0.82\pmstd{0.12}
& \cellcolor{green!0}0.76\pmstd{0.16} \\

w/o procedure queries
& \cellcolor{green!21}0.69\pmstd{0.13}
& \cellcolor{green!22}0.19\pmstd{0.11}
& \cellcolor{green!28}0.84\pmstd{0.11}
& \cellcolor{green!27}0.79\pmstd{0.15} \\

w/o phase queries
& \cellcolor{green!3}0.67\pmstd{0.14}
& \cellcolor{green!0}0.23\pmstd{0.12}
& \cellcolor{green!35}0.85\pmstd{0.10}
& \cellcolor{green!35}0.80\pmstd{0.14} \\

\bottomrule
\end{tabular}}
\end{table}

\subsubsection{Effect of training clip length}
Table~\ref{tab:clip_length} shows how the clip length used during training affects performance. A clip length of 1 corresponds to training without temporal propagation. Increasing the clip length to 5 frames mainly improves temporal consistency, raising mVC12 from 0.82 to 0.85 and mVC24 from 0.76 to 0.79, with a small gain in Dice and HD95. Longer clips bring little further benefit. At 10 and 15 frames, consistency stays at 0.85 mVC12 and gains at most 0.01 mVC24. Over the same range, HD95 worsens from 0.17 to 0.19 and 0.21, and at 15 frames Dice falls back to 0.69. Longer clips may let errors in the propagated queries accumulate over more steps during training, which affects boundary accuracy more than consistency. A clip length of 5 gives the best balance between accuracy and temporal consistency, and we use it for all other experiments.

\begin{table}[h]
\centering
\small
\setlength{\tabcolsep}{3pt}
\caption{Effect of the clip length used for temporal propagation on surgical video segmentation. Values are the mean $\pm$ standard deviation over test clips. Cell shading is scaled linearly within each column from the worst (white) to the best (dark green) value.}
\label{tab:clip_length}
\resizebox{\columnwidth}{!}{%
\begin{tabular}{l | c c c c}
\toprule
Clip length & Dice $\uparrow$ & HD95 $\downarrow$ & mVC12 $\uparrow$ & mVC24 $\uparrow$ \\
\midrule
1 (no temporal prop.)
& \cellcolor{green!4}0.69\pmstd{0.13}
& \cellcolor{green!23}0.18\pmstd{0.10}
& \cellcolor{green!0}0.82\pmstd{0.12}
& \cellcolor{green!0}0.76\pmstd{0.16} \\

5 (ATLAS)
& \cellcolor{green!34}0.70\pmstd{0.13}
& \cellcolor{green!35}0.17\pmstd{0.11}
& \cellcolor{green!29}0.85\pmstd{0.10}
& \cellcolor{green!26}0.79\pmstd{0.15} \\

10
& \cellcolor{green!35}0.70\pmstd{0.13}
& \cellcolor{green!23}0.19\pmstd{0.10}
& \cellcolor{green!35}0.85\pmstd{0.10}
& \cellcolor{green!35}0.80\pmstd{0.14} \\

15
& \cellcolor{green!0}0.69\pmstd{0.14}
& \cellcolor{green!0}0.21\pmstd{0.12}
& \cellcolor{green!35}0.85\pmstd{0.10}
& \cellcolor{green!35}0.80\pmstd{0.14} \\

\bottomrule
\end{tabular}}
\end{table}

\subsubsection{Effect of training data size}
\begin{figure}
    \centering
    \includegraphics[width=0.99\linewidth]{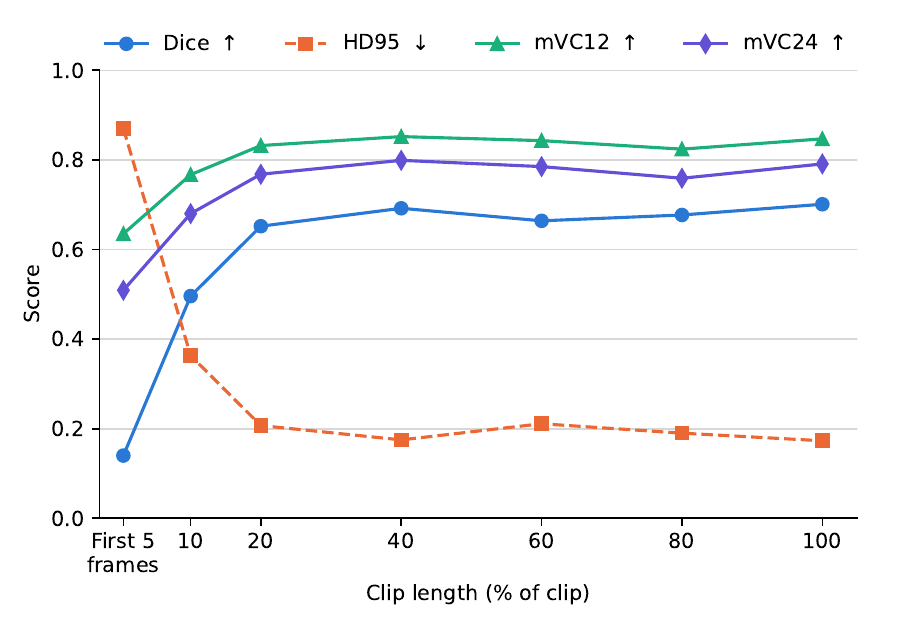}
    \caption{Effect of training data size on segmentation accuracy (mDice, HD95) and temporal consistency (mVC12, mVC24). The x-axis shows the fraction of frames from each training clip used for training and "First 5 frames" corresponds to five annotated frames per clip.}
    \label{fig:data_size}
\end{figure}

Figure~\ref{fig:data_size} shows how performance changes with the fraction of each training clip used for training. With five annotated frames per clip, the model fails to learn useful segmentations: Dice is close to zero and HD95 close to its maximum. Its consistency scores are still moderate (0.63 mVC12), which again shows that consistency cannot be judged separately from accuracy. Performance rises quickly as more of each clip is used. With 10\% of each clip, Dice reaches 0.50, and with 20\% it reaches 0.65, with HD95 dropping from 0.36 to 0.21. From 40\% onwards the curves level off. Dice stays between 0.66 and 0.70, HD95 between 0.17 and 0.21, and mVC24 between 0.76 and 0.80. Training on 40\% of each clip gives results close to the full dataset. This suggests that, beyond this point, extra frames from the same clips add little new information, because consecutive frames are highly redundant.

\subsection{Generalization}
\subsubsection{Zero-shot transfer}
Table~\ref{tab:zero_shot_cholec} compares ATLAS applied zero-shot to CholecSeg8k~\citep{Cholecseg8k}, without any training on CholecSeg8k data, with the same model trained from SurgeNet initialization on the CholecSeg8k training set. The zero-shot model reaches the same Dice (0.83) and a slightly lower HD95 (0.12 vs.\ 0.13). Temporal consistency is slightly lower than for the model trained on CholecSeg8k (0.94 vs.\ 0.96 mVC12 and 0.92 vs.\ 0.95 mVC24). A model trained on ATLAS-120k thus matches the segmentation accuracy of a dedicated model on an unseen dataset, at a small cost in temporal consistency.

\begin{table}[h]
\centering
\small
\setlength{\tabcolsep}{3pt}
\caption{Zero-shot transfer of ATLAS to CholecSeg8k compared with training with SurgeNet initialization on CholecSeg8k.}
\label{tab:zero_shot_cholec}
\begin{tabular}{l | c c c c}
\toprule
Setup & Dice $\uparrow$ & HD95 $\downarrow$ & mVC12 $\uparrow$ & mVC24 $\uparrow$ \\
\midrule
Zero-shot    & 0.83\pmstd{0.08} & 0.12\pmstd{0.06} & 0.94\pmstd{0.03} & 0.92\pmstd{0.03} \\
SurgeNet & 0.83\pmstd{0.10} & 0.13\pmstd{0.08} & 0.96\pmstd{0.02} & 0.95\pmstd{0.02} \\
\bottomrule
\end{tabular}
\end{table}

\subsubsection{Transfer to surgical phase recognition}
Table~\ref{tab:phase_recognition_generalization} reports phase recognition on AutoLaparo when the model is initialized from SurgeNet or from ATLAS weights. ATLAS pretraining improves all metrics. Accuracy increases from 0.82 to 0.85, F1 by 0.03, and balanced accuracy and Jaccard by 0.04.

\begin{table}[h]
\centering
\small
\setlength{\tabcolsep}{3pt}
\caption{Surgical phase recognition on AutoLaparo when initializing from SurgeNet versus from ATLAS pretraining.}
\label{tab:phase_recognition_generalization}
\resizebox{\columnwidth}{!}{%
\begin{tabular}{l l | c c c c}
\toprule
Dataset & Init & Acc $\uparrow$ & Bal. Acc $\uparrow$ & F1 $\uparrow$ & Jaccard $\uparrow$ \\
\midrule
AutoLaparo & SurgeNet & 0.82\pmstd{0.09} & 0.73\pmstd{0.07} & 0.73\pmstd{0.08} & 0.63\pmstd{0.09} \\
           & ATLAS   & 0.85\pmstd{0.07} & 0.77\pmstd{0.08} & 0.76\pmstd{0.09} & 0.67\pmstd{0.10} \\
\bottomrule
\end{tabular}}
\end{table}

\section{Discussion}

\subsection{Main findings}
This work introduced ATLAS-120k, a multi-procedure dataset for surgical anatomy segmentation, and ATLAS, a video segmentation model that conditions its predictions on procedure and phase context. ATLAS outperformed all compared methods on both segmentation accuracy and temporal consistency. The gains did not depend on model size alone: ATLAS ViT-B outperformed SAM3-UNet with about a fifth of the parameters, and at matched backbone size ATLAS ViT-B improved Dice by 0.10 over EndoViT. The ablations show that each component addresses a different part of the problem. In-domain pretraining improved all metrics, query propagation mainly improved temporal consistency, and the phase queries mainly improved accuracy and boundary placement.

\subsection{Role of procedural context}
The phase queries had the largest effect on HD95 of all components, while leaving temporal consistency unchanged. This fits the motivation for the model. In a dataset with 14 procedures, which structures can appear in a frame depends strongly on what is being done and how far the operation has progressed. Context reduces this uncertainty, and the effect is most visible in HD95, which penalizes predictions of structures that are absent and failures to detect structures that are present. The phase queries are trained without phase labels, so this benefit does not require additional annotation and can be extended to procedures for which no phase definitions exist.

The procedure queries contributed less. One explanation is that the procedure is relatively easy to infer from the visual content, so the segmentation queries may already capture much of this information through the patch tokens. In clinical practice the procedure is known before surgery, so it could also be supplied as an input rather than predicted. We kept it as a learned query so that the model does not depend on external metadata, but providing it directly may be a simpler alternative in deployment.

\subsection{Temporal consistency}
Query propagation was the main source of temporal consistency, but its benefit was limited to short training clips. Training on more than five frames did not further improve consistency and worsened HD95, possibly because errors in the propagated queries accumulate over more steps during training. The comparison with the annotators points in the same direction: the gap was 0.11 for mVC$_{12}$ but 0.15 for mVC$_{24}$. Propagating queries from only the previous frame thus keeps predictions stable over short horizons, but not over longer ones. A memory that spans more frames, or that is updated selectively based on prediction confidence, may close part of this gap.

\subsection{Pretraining and architecture}
In-domain pretraining was beneficial but not sufficient. EndoFM and GastroNet5M, both pretrained on endoscopic data, performed worse than DINOv3, and EndoViT and the SurgeNet models were the strongest baselines. The pretraining data and objective therefore matter more than whether the data are endoscopic. For DINOv2 and DINOv3, scaling from ViT-B to ViT-L did not help, which suggests that a linear head limits how much of the encoder's representation can be used for dense prediction. ATLAS avoids a separate decoder as well, but uses the encoder itself to produce masks through the segmentation queries.

At the smallest scale, the SurgeNet models outperformed ATLAS ViT-S. These models use hierarchical backbones with multi-scale features and a dedicated decoder, which may matter more when the encoder has little capacity. The per-class results point to a related limitation. The lowest Dice was found for thin structures such as nerves, arteries, ducts and airways. With an input resolution of $256 \times 256$ and a patch size of 16, each patch covers a region that can be wider than these structures. Higher input resolutions or multi-scale features may therefore be needed for the structures that matter most for surgical safety.

\subsection{Dataset and annotation}
The inter-observer study indicates both how reliable the annotations are and what level of performance can be expected on ATLAS-120k. The annotators agreed with a Dice of 0.85 and were highly consistent over time (0.96 mVC$_{12}$), so the annotations are reproducible, although not exactly. The remaining disagreement likely reflects structures whose extent is ambiguous in the endoscopic view, such as fat and connective tissue. ATLAS ViT-L remained 0.15 Dice below the annotators, leaving substantial room for improvement. The annotation process also gives practical guidance for future datasets. Fine-tuning the tracker on 25\% of the data gave almost all of the improvement in propagation quality, and training ATLAS on 40\% of each clip gave results close to using all frames. Because consecutive frames are highly redundant, annotating more clips sparsely is likely a more efficient use of annotation effort than annotating fewer clips densely.

The standard deviations in Table~\ref{tab:sota_results} were large for all models (0.12--0.16 Dice) and of similar size across models. They reflect differences between test clips rather than uncertainty in the mean. ATLAS-120k covers 14 procedures, and clips differ in which structures are visible and in how large and how clearly visible these structures are. Clips dominated by thin or rare structures are likely harder for every model than clips dominated by large organs, which is consistent with the per-class results. Mean scores on ATLAS-120k should therefore be read together with the per-class results, and differences between models are better judged with paired comparisons over the same clips than from the spread of each model alone.

\subsection{Generalization}
Applied zero-shot to CholecSeg8k, ATLAS matched the Dice of a model trained on that dataset. This suggests that a broad multi-procedure dataset can replace procedure-specific training for at least some procedures. Cholecystectomy is, however, the procedure with the most data in ATLAS-120k, so zero-shot performance on less represented procedures is likely lower. The improvement in phase recognition on AutoLaparo indicates that training on ATLAS-120k produces features that also carry information about surgical workflow, not only about anatomy.

\subsection{Limitations}
This study has several limitations. All videos were taken from a single source, the GSViT dataset, and come with little information on patients, centers or recording equipment. Performance on other institutions and systems therefore remains to be confirmed. The class distribution is long-tailed, and several rare classes occur in only a few test clips, so their per-class results are uncertain. Some clinically distinct structures were merged into broader classes because they are hard to tell apart in the endoscopic view, which limits the level of anatomical detail the model provides. Most clips are shorter than 30 seconds, so temporal consistency was evaluated only over short horizons and not over full procedures. Finally, segmentation quality was evaluated with technical metrics only; whether the predictions are accurate enough to support surgeons during an operation was not assessed.

\section{Conclusion}
We introduced ATLAS-120k, a dataset of 121,018 densely annotated frames covering 14 laparoscopic and robot-assisted procedures, and ATLAS, a video segmentation model that combines procedure and phase context queries with query propagation between frames. ATLAS outperformed natural-image and in-domain foundation models on both segmentation accuracy and temporal consistency, matched a dedicated model zero-shot on CholecSeg8k, and runs well above real-time frame rates. It still falls 0.15 Dice short of inter-observer agreement, and thin structures such as nerves, arteries and ducts remain the hardest to segment. We release the ATLAS-120k annotations and the ATLAS model weights to support further work toward reliable anatomy recognition during surgery.

\printcredits

\section*{Declaration of competing interest}
Ronald L. P. D. de Jong reports financial support was provided by Nederlandse Organisatie voor Wetenschappelijk Onderzoek Toegepaste en Technische wetenschappen. Yiping Li reports financial support was provided by Hanarth Fund Foundation. Jelle P. Ruurda reports financial support was provided by Intuitive Surgical Inc. Jelle P. Ruurda reports financial support was provided by Medtronic Inc. Jelle P. Ruurda reports financial support was provided by Olympus Corporation. Josien P. W. Pluim is one of the editors of the Medical Image Analysis Journal. If there are other authors, they declare that they have no known competing financial interests or personal relationships that could have appeared to influence the work reported in this paper.

\section*{Data availability}
The ATLAS-120k annotations, model weights and annotation tool are publicly available at \url{https://github.com/TimJaspers0801/ATLAS}. The source videos are part of the GSViT dataset \citep{schmidgall2024general}.

\section*{Acknowledgments}
This research was funded by Stichting Hanarth Fonds, study number 2022-13, and the Dutch Research Council (NWO), study number KICH1.ST03.21.019, grant ID https://doi.org/10.61686/NKMZR04566. It is part of the INTRA-SURGE (INTelligent computeR-Aided Surgical gUidance for Robot-assisted surGEry) project aimed at advancing the future of surgery. Furthermore, we thank SURF (www.surf.nl) for the support in using the Dutch National supercomputer Snellius.

\section*{Declaration of generative AI and AI-assisted technologies in the manuscript preparation process}
During the preparation of this work the authors used Claude (Anthropic) to improve the readability and language of the manuscript and to assist in creating the schematic in Figure 4. After using this tool, the authors reviewed and edited the content as needed and take full responsibility for the content of the publication.

%\bibliographystyle{unsrtnat} % numeric, works with natbib
\bibliographystyle{cas-model2-names} % author–year (names) style

\bibliography{cas-refs}

\newpage
\appendix
\balance
\section{Dataset splits}\label{app:splits}
Table~\ref{tab:split_classes} shows how the annotated classes are distributed over the training, validation and test splits. The splits contain 70, 10 and 20 videos, respectively, with 368, 39 and 95 clips and 84,113, 9,604 and 27,301 annotated frames. All 14 procedures occur in the training set, whereas the validation and test sets cover 6 and 9 procedures. As a result, the class distributions differ between splits. The most frequent classes, such as tools/camera, fat, abdominal wall and liver, occur in all splits. Classes related to cholecystectomy, such as gallbladder and cystic duct, are more common in the validation and test sets than in training. Most procedure-specific classes, for example those of prostatectomy, occur in only one or two test videos. All evaluated classes are present in the training set.

\begin{table*}[h]
\centering
\footnotesize
\setlength{\tabcolsep}{4pt}
\caption{Class distribution over the training, validation and test splits of ATLAS-120k, sorted by the total number of annotated frames. For each split, the percentage of annotated frames in which the class occurs and the number of videos that contain it are given. Classes that were merged into a parent class for training and evaluation list that parent class in the second column; classes marked -- were relabeled as background.}
\label{tab:split_classes}
\begin{tabular}{l l | r r | r r | r r}
\toprule
 & & \multicolumn{2}{c|}{Train} & \multicolumn{2}{c|}{Val} & \multicolumn{2}{c}{Test} \\
Annotated class & Evaluated as & \% frames & Videos & \% frames & Videos & \% frames & Videos \\
\midrule
Tools/camera &  & 97.9 & 70 & 98.6 & 10 & 99.1 & 20 \\
Omentum & Fat & 41.4 & 38 & 60.0 & 7 & 57.4 & 14 \\
Abdominal wall &  & 42.2 & 43 & 42.2 & 8 & 48.4 & 12 \\
Liver &  & 31.3 & 29 & 56.8 & 7 & 50.8 & 12 \\
Colon/rectum &  & 32.4 & 26 & 11.4 & 3 & 27.3 & 8 \\
Non-anatomical &  & 28.8 & 35 & 16.6 & 6 & 18.7 & 14 \\
Gallbladder &  & 17.1 & 24 & 46.1 & 5 & 44.1 & 10 \\
Small intestine &  & 29.9 & 27 & 14.8 & 6 & 8.4 & 7 \\
Stomach &  & 18.7 & 15 & 17.3 & 3 & 16.1 & 8 \\
Mesentery & Fat & 20.1 & 14 & 7.1 & 2 & 6.6 & 2 \\
Artery &  & 16.0 & 19 & 14.6 & 3 & 14.8 & 12 \\
Spleen &  & 6.4 & 4 & 15.2 & 1 & 6.3 & 1 \\
Cystic duct & Bile duct & 3.2 & 10 & 21.7 & 4 & 13.6 & 8 \\
Vein &  & 9.2 & 23 & $<$0.1 & 1 & 1.0 & 5 \\
Azygos vein & Vein & 5.5 & 8 & 11.5 & 1 & 5.6 & 2 \\
Lung &  & 4.8 & 8 & 10.3 & 1 & 5.8 & 2 \\
Prostate &  & 3.6 & 6 & 2.1 & 1 & 8.8 & 2 \\
Duodenum & -- & 5.9 & 4 & -- & -- & -- & -- \\
Seminal vesicle &  & 3.1 & 4 & -- & -- & 4.6 & 1 \\
Pericardium &  & 1.6 & 6 & 9.6 & 1 & 5.2 & 2 \\
Hepatic ligament &  & 0.9 & 5 & 13.1 & 4 & 5.9 & 4 \\
Mesocolon & -- & 4.3 & 4 & $<$0.1 & 1 & -- & -- \\
Pancreas & -- & 4.2 & 2 & -- & -- & -- & -- \\
Urethra &  & 2.1 & 3 & 4.8 & 1 & 3.2 & 2 \\
Fallopian tubes &  & 2.0 & 3 & 1.4 & 1 & 2.9 & 1 \\
Ligated plexus &  & 2.0 & 3 & 3.1 & 1 & 1.9 & 1 \\
Ovary &  & 2.0 & 3 & -- & -- & 2.6 & 1 \\
Major nerve &  & 2.1 & 6 & -- & -- & 1.5 & 1 \\
Uterus &  & 1.6 & 3 & 1.4 & 1 & 2.5 & 1 \\
Bladder &  & 1.3 & 4 & 3.8 & 1 & 2.5 & 2 \\
Cystic plate &  & 1.1 & 5 & 3.1 & 3 & 3.3 & 4 \\
Diaphragm & -- & -- & -- & 7.0 & 4 & 5.2 & 5 \\
Trachea/bronchus &  & 1.5 & 5 & 1.6 & 1 & 1.9 & 1 \\
Aorta & Artery & 1.8 & 4 & -- & -- & $<$0.1 & 1 \\
Vena cava & Vein & 1.5 & 3 & -- & -- & -- & -- \\
Catheter & Non-anatomical & 0.8 & 3 & 1.9 & 1 & 1.1 & 1 \\
Adrenal gland & -- & 0.9 & 1 & -- & -- & -- & -- \\
Common bile duct & Bile duct & 0.7 & 2 & -- & -- & 0.5 & 1 \\
Kidney & -- & 0.5 & 2 & -- & -- & -- & -- \\
Esophagus & -- & 0.4 & 2 & -- & -- & 0.2 & 1 \\
Hepatic duct & Bile duct & 0.5 & 1 & -- & -- & -- & -- \\
\bottomrule
\end{tabular}
\end{table*}

\end{document}